\documentclass[10pt,twocolumn,letterpaper]{article}

\usepackage{wacv}
\usepackage{times}
\usepackage{epsfig}
\usepackage{graphicx}
\usepackage{amsmath}
\usepackage{amssymb}
\usepackage{booktabs}
\usepackage{float}
\usepackage[utf8]{inputenc} 
\usepackage[T1]{fontenc}
\usepackage{url}
\usepackage{amsfonts}
\usepackage{nicefrac}
\usepackage{microtype}
\usepackage{xcolor}
\usepackage{times}
\usepackage{graphicx}
\usepackage{enumitem}
\usepackage{adjustbox}
\usepackage{latexsym}
\usepackage{tabularx}

\wacvapplicationstrack 

\wacvfinalcopy 

\ifwacvfinal
\usepackage[breaklinks=true,bookmarks=false]{hyperref}
\else
\usepackage[pagebackref=true,breaklinks=true,colorlinks,bookmarks=false]{hyperref}
\fi

\begin{document}

\title{SAVCHOI: Detecting \underline{S}uspicious \underline{A}ctivities using Dense \underline{V}ideo \underline{C}aptioning with \underline{H}uman \underline{O}bject \underline{I}nteractions 
}

\author{Ansh Mittal \thanks{Author contributed majorly to this research}\\
  Department of Computer Science\\
  University of Southern California\\
  Los Angeles, CA 90007 \\
  {\tt\small anshm@usc.edu}
 \and Shuvam Ghosal\\
  Department of Computer Science\\
  University of Southern California\\
  Los Angeles, CA 90007 \\
  {\tt\small sghosal@usc.edu}
 \and Rishibha Bansal\\
  Department of Computer Science\\
  University of Southern California\\
  Los Angeles, CA 90007 \\
  {\tt\small bansalr@usc.edu}}

\maketitle
\thispagestyle{empty}

\begin{abstract}
Detecting suspicious activities in surveillance videos is a longstanding problem in real-time surveillance that leads to difficulties in detecting crimes. Hence, we propose a novel approach for detecting and summarizing suspicious activities in surveillance videos. We have also created ground truth summaries for the UCF-Crime video dataset. We modify a pre-existing approach for this task by leveraging the Human-Object Interaction (HOI) model for the Visual features in the Bi-Modal Transformer. Further, we validate our approach against the existing state-of-the-art algorithms for the Dense Video Captioning task for the ActivityNet Captions dataset. We observe that this formulation for Dense Captioning performs significantly better than other discussed BMT-based approaches for BLEU@1, BLEU@2, BLEU@3, BLEU@4, and METEOR. We further perform a comparative analysis of the dataset and the model to report the findings based on different NMS thresholds (searched using Genetic Algorithms). Here, our formulation outperforms all the models for BLEU@1, BLEU@2, BLEU@3, and most models for BLEU@4 and METEOR falling short of only ADV-INF Global by 25\% and 0.5\%, respectively.
\end{abstract}

\section{Introduction}
\label{sec:1}
We are interested in the problem of detecting suspicious (or criminal) activities from a real-time surveillance feed. For this purpose, several approaches have been discussed \cite{ullah2021efficient, maqsood2021anomaly, majhi2021weakly}, but not many formulations can detect criminal activities in surveillance feeds. Today, there are CCTV cameras at every corner, commercial outlet, and school. Hence, it's challenging to process and analyze this large amount of surveillance feed by humans. Monitoring this data remains a challenging problem due to issues like restricted access to surveillance data because of privacy concerns and poor quality of surveillance footage.

\begin{figure*}[!ht]
    \centering
    \includegraphics[width=\textwidth]{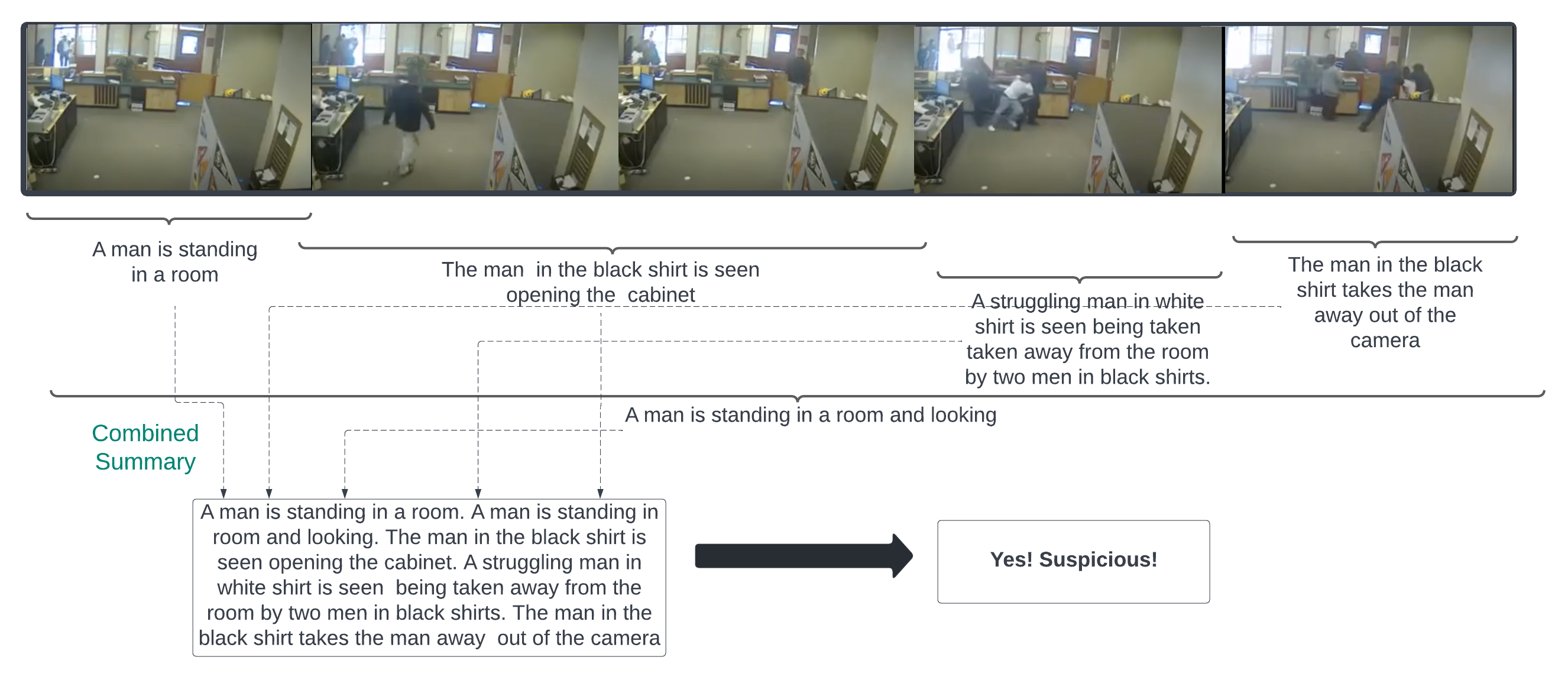}
    \caption{An end-to-end working example with the proposed approach. The information is not repeated several times due to Non-Maximum Suppression (NMS). The NMS threshold is 0.47 (value obtained after using genetic optimizations; discussed in detail later).}
    \label{fig:example}
\end{figure*}

\begin{figure*}[!htp]
    \centering
    \includegraphics[width=\textwidth]{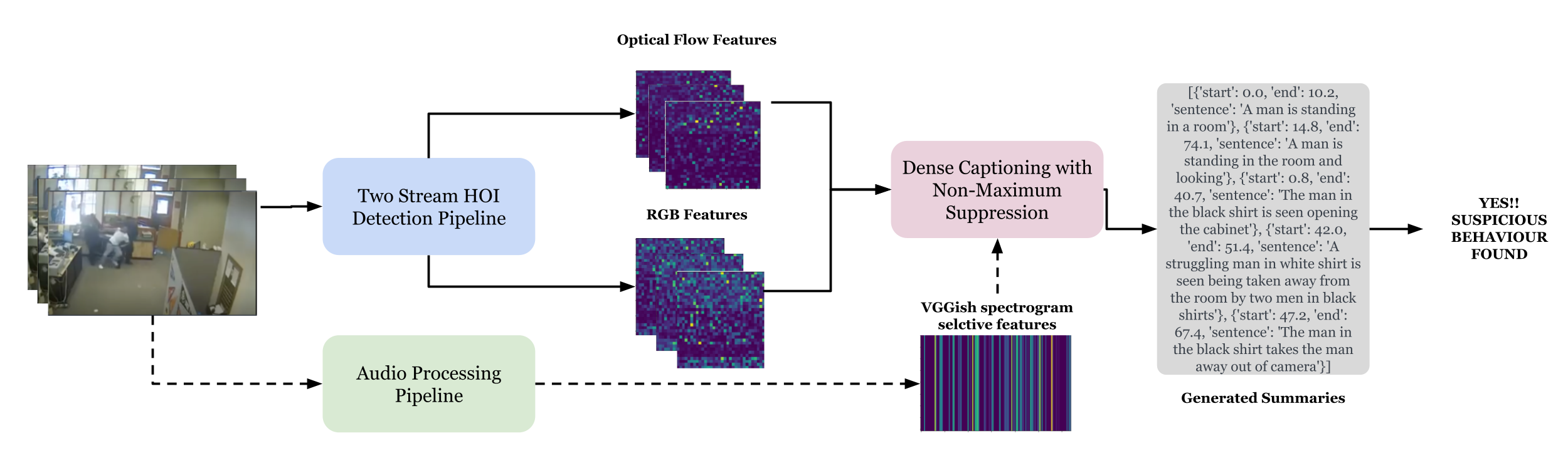}
    \caption{The proposed data flow is based on the proposed architecture defined in section \ref{sec:proposedmethodsandImplementation}}
    \label{fig:workflow}
\end{figure*}

This research presents a novel approach for detecting suspicious and criminal behavior in videos by leveraging and combining the architecture of two different algorithms that are state-of-the-art in detecting Human-Object Interactions and Dense Video Captioning. This paper discusses the problem formulation as a combination of dense summary generation (using a two-way disentanglement approach) in the videos and then classifying the generated summaries. Hence, the architecture uses a sequential combination of a Human-Object Interaction pipeline and selective Bi-Modal Dense Video Captioning (for Audio inputs). This paper leverages the downstream task of text classification to emphasize the capability of the summary generated. For this, the classifier was trained exhaustively on crime-based data to classify the captions as suspicious or not (criminal and non-criminal). This architecture allows us to transfer the task of crime detection using video (for which the data is not freely accessible) to crime detection using text (an easily accessible modality). Figure \ref{fig:example} presents a working example of the proposed method. This sequence of images is from a video of the `Arrest' category in the UCF-Crime dataset. Figure~\ref{fig:workflow} depicts the data flow diagram for our approach.

This work proposes a five-fold contribution to the pre-existing literature for dense captioning and surveillance-based detection.
\begin{enumerate}
\item Introduced a novel modified architecture for detecting anomalous and suspicious behaviors in videos.
\item Augment ground-truth summaries to the UCF-crime video dataset~\cite{sultani2018real} for training the modified models.
\item Curate a dataset for the text-based description of suspicious and non-suspicious activities using the CHARADES dataset~\cite{sigurdsson2016hollywood} and the DPS report dataset\footnote{https://dps.usc.edu/category/alerts/} (which was scraped using BeautifulSoup).
\item Presented optimization setting of a hyperparameter (here, NMS threshold value) of the model to improve the BLEU@1 of the generated summaries.
\item Perform comparative analysis on proposed model architectures and evaluate them on the ActivityNet Captions dataset.
\end{enumerate}

The remainder of this research work is structured as follows. Section~\ref{relworks} discusses a literature survey on the problem and architecture. Section~\ref{data} discusses the datasets used and Section~\ref{sec:proposedmethodsandImplementation} expands on the proposed architecture. The results are discussed in Section~\ref{results} and further analyzed and compared in Section~\ref{ablative}. We discuss some Dataset Limitations and Ethical Considerations in~\ref{sec:datasetlimitations} and~\ref{sec:ethicalconsiderations}, respectively. Finally, we conclude this research in Section~\ref{conclusion} with a possibility for future extensions to this work. The final section then discusses the ethical considerations entailed in this research.

\section{Related Work}

\label{relworks}
This section points to the existing literature on Human-Object Interactions in conjunction with Video Human Activity Recognition and Dense Captioning. Hence, forming the basis of understanding our research. Further, this section later discusses works related to real-time surveillance-based crime detection.

\subsection{Human-Object-Interaction Recognition}

The transformer architecture (or encoder-decoder architecture) works well for Action Recognition and Action Detection in videos. This idea was initially explored by~\cite{ji2019video} and became more prominent with the the recent advent of Vision Transformers~\cite{dosovitskiy2020image} and subsequent works in the field of videos: STAM~\cite{sharir2021image}. The contribution made by \cite{girdhar2019video} discusses tasks such as action detection using the AVA dataset \cite{gu2018ava} by disentangling the human hands from faces. Detection Transformer (DE-TR) \cite{carion2020end} proposes a streamlined approach for leveraging set-based global loss that forces unique predictions via bipartite matching using a transformer encoder-decoder architecture. Later works have introduced architectures like CDN~\cite{zhang2021mining} based on a disentangling paradigm for Human-Object Interaction tasks trained on the HICO-DET dataset. This architecture disentangles the originally set-based optimization functions into two different cascaded decoders for human-object detection and interaction recognition. Unary Pairwise Transformer (UPT) \cite{zhang2022efficient} (based on the DE-TR model and pre-trained on the HICO-DET and V-COCO datasets) tries to understand the pairwise interactions between humans and objects. The recently proposed and state-of-the-art model QAHOI~\cite{chen2021qahoi} (Query-based Anchors for Human Object Interaction Detection)--based on a SWIN-transformer~\cite{liu2021swin} as its backbone ensures that its detections are spatially coherent. Meanwhile, the querying mechanism ensures that only the interaction with the highest confidence score gets chosen between different Human and Object interactions. A more recent approach to action recognition deals with the usage of Timesformer \cite{bertasius2021space}, where self-attention is applied temporally and spatially within each patch on datasets like Kinetics-600 \cite{carreira2018short}.

\begin{table*}
\caption{A brief Comparison of detection made by CDN, QAHOI, and fine-tuned DE-TR models.}
\label{tab:comparetable}
\centering
\begin{tabularx}{0.85\linewidth}{c|c|c|c|c|c|c}
\toprule \textbf{Model} & \textbf{Full(D)} & \textbf{Rare(D)} & \textbf{Non-Rare(D)} & \textbf{Full(KO)} & \textbf{Rare(KO)} & \textbf{Non-Rare(KO)}\\ 
\midrule
UPT-base & 32.04 & 29.21 & 33.71 & 34.89 & 29.81 & 36.49\\
CDN-S (R50) & 31.44 & 27.39 & 32.64 & 34.09 & 29.63 & 35.42\\
CDN-B (R50) & 31.78 & 27.55 & 33.055 & 34.53 & 27.93 & 35.96\\
CDN-L (R50) & 32.07 & 27.19 & 33.53 & 34.79 & 29.48 & 36.38\\
QAHOI-Tiny & 28.41 & 22.47 & 30.27 & 30.99 & 24.83 & 32.84 \\
QAHOI-Base+ & 33.58 & 25.86 & 35.88 & 35.34 & 27.24 & 37.76\\
QAHOI-Large+ & \textbf{35.78} & \textbf{29.80} & \textbf{37.56} & \textbf{37.59} & \textbf{31.36} & \textbf{39.36}\\
DE-TR (fine-tuned) & 31.56 & 28.41 & 29.23 & 35.41 & 30.71 & 37.52 \\
\bottomrule
\end{tabularx}
\end{table*}

\subsection{Surveillance-based Crime Detection}
Currently, human activity detection in surveillance videos relates closely to formulating the problem of anomaly detection as a binary classification task \cite{sabokrou2018adversarially, sabokrou2017deep, sultani2018real}. In this domain, Action Recognition and Detection have become vital for detecting criminal activities like robbery, shooting, fighting, etc. Moreover, a substantial amount of research postulates this problem of Surveillance-based Crime Detection as a multi-task problem with multiple modalities~\cite{ullah2021efficient, maqsood2021anomaly, majhi2021weakly}. It suggests that advancement in deep learning is one of the primary motivations for advancing Human Activity Recognition in video surveillance data. Feature extraction \cite{sultani2018real, wu2021weakly, majhi2021weakly, wan2021anomaly} has played a vital role in detecting and recognizing Spatio-temporal features using 3D Convolutions \cite{tran2015learning, carreira2017quo}. Recently, two different techniques--namely, Attention mechanisms \cite{wu2021weakly, majhi2021weakly, ullah2021efficient} and Graph Representation Learning \cite{zhong2019graph, cao2022adaptive}--have been used in surveillance videos for detecting various features and have been one of the motivations for choosing an attention-based mechanism in this contribution. Despite leveraging various techniques, the application of crime-based video detection remains a challenge to be solved efficiently. It is primarily due to the technical and other challenges discussed in the next sub-section. 

\subsection{Dense Video Captioning}
This task was first defined by \cite{krishna2017dense}, and it later integrated context awareness using Bidirectional Single-Stream Temporal (SST) Action Proposals in \cite{buch2017sst} and Self-Critical Sequence Training using Actor-Critic Method with different metrics (CIDEr and METEOR \cite{rennie2017self}). In 2019, \cite{mun2019streamlined, iashin2020multi, iashin2020better} implemented coherent captioning using the SST module and Pointer Networks. With the advent of transformer architectures, there has been an increase in end-to-end architectures for dense captioning \cite{rahman2019watch, iashin2020better, iashin2020multi}. But, the task of creating video-descriptive summaries for real-time surveillance is yet to be solved effectively. In these applications, the earlier hypotheses got limited by the data availability. The text-descriptive capability of models provides visual-semantic feature representation from video in natural language. This perspective suggests the usage of exhaustive descriptions of videos by partially (or completely) determining the actions (here, in a surveillance video) that need to be summarized. It had been studied in more detail by~\cite{estevam2021tell}, which led to leveraging natural language descriptions while recognizing actions in a zero-shot setting for the UCF-101~\cite{soomro2012ucf101}. Later,~\cite{dilawari2021natural} proposed a method for understanding the video streams and storing them as text (to save on storage space) using natural language generation. We have used convolutions and bi-directional recurrent layers for the same. This approach used action recognition, object recognition, and human-specific datasets like UET Video Surveillance, AGRIINTRUSION, and TRECViD \cite{khan2012natural}.

\section{Datasets}
\label{data}

Different datasets used for training different modules of the architecture--Human Object Interaction module, Dense Video Captioning module, and Text Classification task, are discussed in this section. Further, we also present our own two data-based contributions. 

\begin{enumerate}
\item A dataset for ground truth captions of UCF crimes  \cite{sultani2018real} data(described in section \ref{sec: dense_caption_data}). 
\item A dataset for a text-based description of suspicious and non-suspicious activities (described in section \ref{sec:NLP_data}).
\end{enumerate}

\subsection{Human-Object Interaction (HOI) Recognition}

\textbf{HICO-DET}~\cite{chao2018learning} is a dataset consisting of more than 150,000 human-object pair annotations. We split this dataset into a training set and a test set of 117,871 and 33,405, respectively. Furthermore, it contains 80 classes of objects, 117 types of actions, and 600 interaction types (including no interactions for human activity recognition). HICO-DET is one of the most primarily used datasets in HOI detection tasks. Another such dataset is the \textbf{V-COCO} which contains 2,533 training images, 2,867 for validation, 4,946 test images, and only 24 unique action classes.

\subsection{Video summarization}
\label{sec: dense_caption_data}
\textbf{UCF Crimes} \cite{sultani2018real} is a large-scale dataset that consists of 128 hours of videos. It contains 1900 long and untrimmed real-world surveillance videos which show different criminal activities like Burglary, Arrest, Assault, etc. This dataset mainly has two tasks--detecting anomalies in videos and classifying suspicious behavior (or crime-based activity) from the video into any available action classes. We use a subset of the UCF Crimes dataset with only 300 videos (due to computation limitations) to evaluate the models (depicted in section \ref{sec:eval_vs}). We generated ground-truth summaries of these videos by watching the videos for 32 hours, with the following classes (Arrest, Arson, Assault, Burglary, Explosion, Normal, Road Accidents, Robbery, Shooting, Shoplifting, and Vandalism). We used these to evaluate models and for them to function as a dataset for future applications\footnote{We plan to make this dataset publicly available for future analysis}.

\begin{table*}[ht]
\centering
\begin{tabular}{c|p{0.8\textwidth}}
\toprule
\toprule
\textbf{Video class} & \textbf{Created summaries} \\
\midrule
Normal & Cars are parked on the side of the road. Two men enter a building. An old man passes by.\\
Burglary & A man wearing a helmet comes in front of a door of a house. Then, he tries to unlock the door. After several attempts, he fails to open the door and leaves the area.\\
Assault & A man wearing a white shirt and carrying a bag comes and starts beating a man in blue shirt. Some people come forward to stop it.\\
Shooting & A car is seen standing at the side of a road. Another car comes and a person inside the car shoots at the parked car. After that, the second car leaves and the man inside the parked car comes out.\\
Fighting & Two young men wearing white shirts start fighting with a man in a suit. They hit him with a chair and kick him. The man wearing the suit also fights back. Some other people then join and try to control the fight. Finally, both the parties leave the area.\\
Shoplifting & A group of men enter a shop. They take some items from the shop. Some people inside the shop run away. Then, they leave the shop one by one.\\
Vandalism & A car parks on the road. Two men walk toward the car. One men stops next to the car. The man kicks the car. Two men continue walking.\\
Arson & A man is seen in a garage. He sets one of the cars on fire and runs away. Multiple cars catch fire and start burning. The cars are severely damaged.\\
Robbery & A man wearing a white hat and black shirt comes in and aims at an employee at the cash counter. The employee gives away the entire cash to the man.\\
Arrest & Two men come in a bike and one of them tries to get inside a building. Some officers stop him and the man starts fighting them. Finally, the officers arrest him.\\
\bottomrule
\bottomrule
\vspace{1ex}
\end{tabular}
\caption{A sample of author-created ground truth summaries for the UCF-Crime dataset}
\label{gen-summary-table}
\end{table*}

\subsection{Text classification}
\label{sec:NLP_data}
A Text-based classifier classifies the video captions generated. The dataset for training the NLP module has descriptions of crime scenes in the present tense (from the third person view) and non-criminal activity descriptive summaries. For criminal activities, the `incident description' section of \textbf{DPS crime alerts} has been scraped from the DPS website\footnote{\url{https://dps.usc.edu/category/alerts/} (accessed: May 8, 2022)} using the BeautifulSoup module~\cite{richardson2007beautiful}, a total of 212 reports. We augmented the generated data to include multiple permutations of the words--`man' and `woman' instead of `suspect' and `victim.'

For non-criminal activities, we extracted 7985 transcripts from the \textbf{CHARADES} dataset~\cite{sigurdsson2016hollywood}, where each transcript is one sentence long and describes a human activity. We merged three transcripts to represent one activity to make the length comparable to summaries in the suspicious class, resulting in 2661 activities. This merge does not cause a problem because all activities of the CHARADES dataset can occur in conjunction with each other (without inconsistencies) as they are all individual activities. Moreover, we converted everything to the present tense as our generated ground truth video summaries have present tense\footnote{All activities are taken in the present tense (using the tenseFlow (\url{https://github.com/bendichter/tenseflow} (accessed: May 8, 2022)) to maintain symmetry between both kinds of descriptions. Stop words are removed and all sentences contain either `man' or `woman' and not `a person'.}. We picked approximately 408 summaries from this generated text corpus, and then used them as samples for non-suspicious activities to fine-tune the pre-trained \textbf{BERT-base-uncased} model.

Further, we added 250 criminal activity descriptions (label `1') from the generated GloVe embeddings-based summaries in UCF-Crime  (section \ref{sec: dense_caption_data}) and 50 summaries for non-suspicious activities  (label `0') descriptions to the NLP text corpus for training the classifier. Hence, concluding our text corpus for fine-tuning the NLP text classifier with 920 textual summaries (out of which 445 are crime-based and 475 are normal activities).

\section{Proposed Methodology}
\label{sec:proposedmethodsandImplementation}
\begin{figure*}[!htp]
    \centering
    \includegraphics[width=0.9\textwidth]{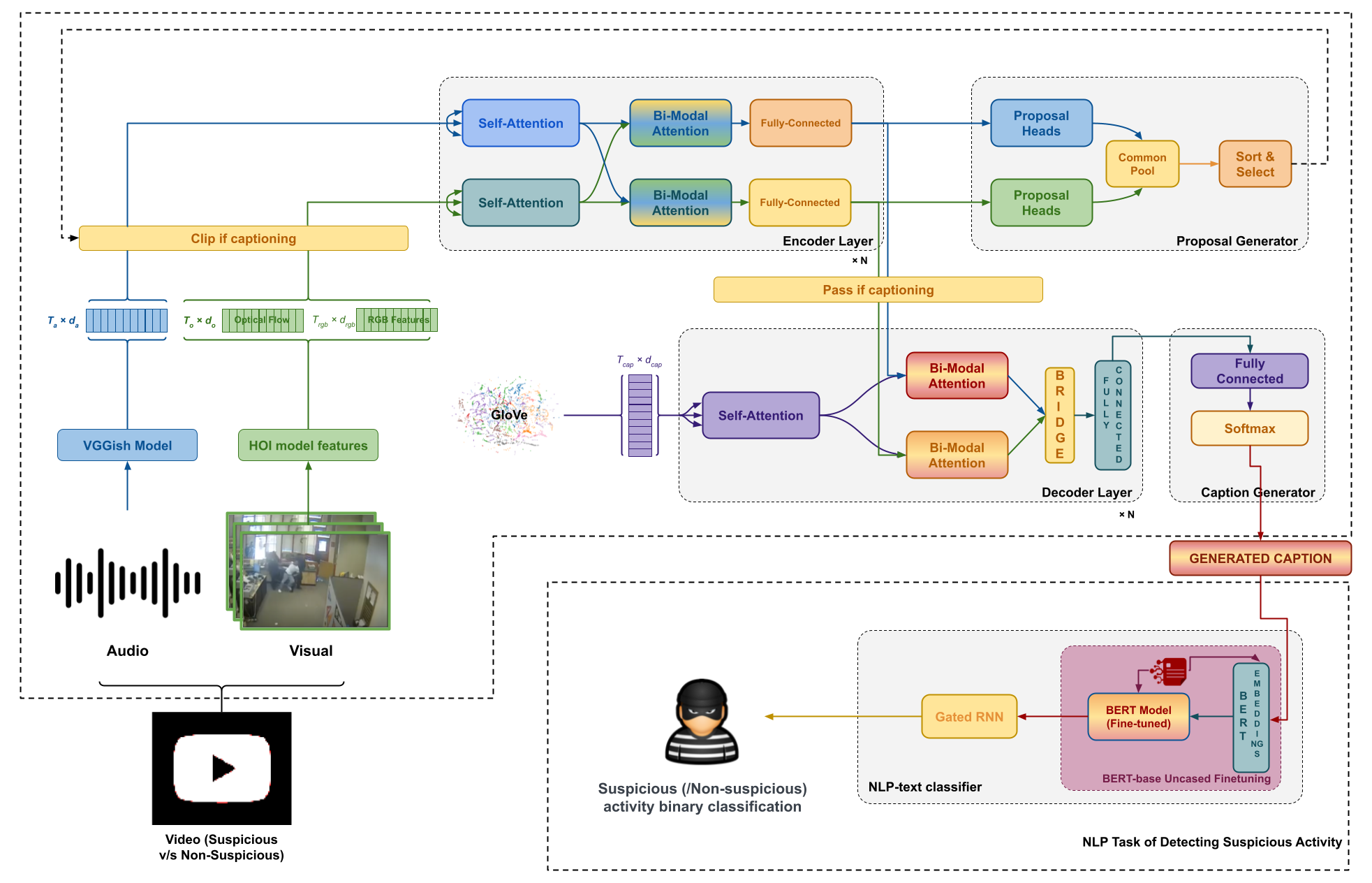}
    \caption{The proposed new architecture for detecting suspicious activities in videos (\textbf{BMT + Seq HOI + NMS}).}
    \label{fig:architecture}
\end{figure*}

Figure~\ref{fig:architecture} and figure~\ref{fig:bmtparhoi} represent the different architectures that this research proposes. Figure~\ref{fig:bmtparhoi} (BMT+Par HOI+NMS) proves to be suboptimal architecture, inferred from the results obtained in tables~\ref{tab:ucf-bleu-table} and~\ref{tab:mdvc-bmt-table}. We discuss this architecture in detail in subsection~\ref{subsec:evalActivityNet}. Instead, we discuss BMT+Seq HOI+NMS (Figure~\ref{fig:architecture}) in much more depth in this section. Here, the NMS. In this model, the Human-Object Interaction module replaces the I3D model features in a Sequential fashion. Further, this model utilizes the concept of Non-Maximum Suppression (NMS; given in more detail in Supplementary~\ref{sec:NMS}) based on the Spatio-temporal dimension of the video.

We support the Sequential architecture as it accumulates the gradients during training and generates Optical Flow (using time-sensitive Intersection-over-Union discussed) and RGB (as depicted in figure~\ref{fig:workflow}) feature outputs. The object query of the Proposal Generator yields the RGB features, which then clip the videos into subsequences of videos to be encoded and decoded by BMT encoder-decoder architecture. Further, the Decoder architecture uses the encoded features from the trained Bi-Modal Attention in the Encoder layer to understand (decode) the feature vectors using a combination of self-Attention layers and GloVe embeddings to the BiModal Attention layer. For this task, we have leveraged several engineering and novel research techniques in this pipeline, as represented in figure~\ref{fig:architecture}. This process of encoding and decoding repeats $N$ times until all the Optical Flow features and the RGB features are encoded and decoded to give a sequence of words, which is the required caption for the video. These are as follows.

\begin{enumerate}[noitemsep]
\item Selective feature engineering for Audio-based VGGish where they were not available (videos without audio are assigned 0-valued tensors) as UCF-Crime is a dataset that is predominantly audio-less;
\item Non-Maximum Suppression thresholding of the time-Intersection-over-Union (tIoU) leveraged from the previous research related to BMT~\cite{iashin2020better};
\item Parallel and Sequential HOI-model trained in parts (training of decoder and encoder of the model occurs separately to each other) using 20 \% HICO-DET dataset (as depicted in figures~\ref{fig:architecture} and \ref{fig:bmtparhoi})
\end{enumerate}

\textbf{Video Summarization} is implemented using BMT (Bi-modal Transformers)~\cite{iashin2020better} by leveraging HOI embeddings (instead of I3D extracted using the PWC-NET model~\cite{carreira2017quo}) and selective VGG encoding for the audio information. As mentioned earlier, we use selective audio tensors because videos may or may not have sounds that lead the model to propagate blank (or zero) tensors with the same shape as would have been sent if there was audio present in the video. This configuration occurs because most surveillance videos do not have any audio component\footnote{This is due to the US Code Title XVIII Section 2510 which concerns user privacy and prevents anyone from recording audio without user consent. It states that ``verbal communication between two people believing that their conversation is not being intercepted is justifiable reason to assume it is not being recorded''}. This audio-less behavior of surveillance feeds was evident in the UCF-crime dataset, where most videos did not contain audio. As the University of Central Florida compiled the UCF-Crime dataset, where the law selectively prohibits recording audio in surveillance feeds, it is easy to figure out why the UCF-Crime has videos without little to no audio component. For evaluation, we created a separate dataset by checking the videos from the subset of the UCF-Crime dataset (sample depicted in Table \ref{caption-table4}).

\textbf{RGB Feature Extraction based on HOI} We tested models like DE-TR, UPT, CDN, and QAHOI for comparative analysis (depicted in table~\ref{tab:comparetable} and figure~\ref{fig:comphoi}). We used DE-TR primarily due to its good mAP (mean-Average Precision; discussed further in~\ref{sec:mAP}) for higher sampling rates. But, it is vital to understand that DE-TR isn't the top state-of-the-art model in terms of mAP. A Sampling rate of around 35 frames/second gives good results for the DE-TR model, whereas the UPT model works best around the sampling rate of 25 frames/sec. Further, models like CDN and QAHOI -- models in the top-5 state-of-the-art for Human-Object-Interaction, sample frames at a rate of about 15 frames/sec, and 10 frames/sec (at best), respectively. These models were evaluated on 2 V100 and 2 A100 GPUs and took about 5 hours 42 minutes and 7 hours 17 minutes, respectively. Hence, due to the low sampling rate of the QAHOI, CDN, and UPT models, we trained the comparatively faster DE-TR for 1000 epochs for 20\% of the HICO-DET dataset. This model training required us \~52 hours on 2 A100 GPUs. This model was later used to generate the flow files and RGB files (similar to the i3D model in the BMT architecture) to ensure the consistency of the model.

Due to issues with the summaries generated from the Vanilla BMT model~\cite{iashin2020better} (Table \ref{caption-table4}), the pipeline suffers from incorrect classifications (False Positives and False Negatives). These false detections can cause problems in monitoring surveillance videos for criminal activities. Hence, we proposed earlier to use transformer-based architecture for recognizing Human-Object triplets. This HOI model replacement ensures that summaries capture the feature representations of the Human-Object-Interactions that the following works--Multi-modal Dense Video Captioning model \cite{iashin2020multi} and the Bi-Modal Transformer~\cite{iashin2020better}, fail to capture. Then text-generation models like \cite{radford2019language} are used in the Decoder Layer to generate a summary from the captions obtained with timestamps. This approach increases the evaluation score of our model on various metrics such as BLEU@1, BLEU@2, BLEU@3, BLEU@4, and METEOR.

\textbf{Text classification:} We fine-tune a pre-trained BERT model~\cite{devlin-etal-2019-bert} on the activity dataset (described in section~\ref{sec:NLP_data}). We used the BMT-generated summaries for inference in the activity dataset, which comprises approximately 920 (out of which 445 are crime-based and 475 are normal activities). We used the \texttt{‘BERT-based-uncased’} model from the Transformers library\footnote{\url{https://github.com/huggingface/transformers} (Accessed: May 8, 2022)} for the model and embedding layer (depicted in figure~\ref{fig:architecture} and figure~\ref{fig:bmtparhoi}). The representations are fed into a Gated-Recurrent Unit (GRU) for performing text classification of criminal activity descriptions. The task of the model is to classify the activity We train this Bi-directional Gated BERT for two epochs with four layers, 256 hidden dimensions, 15\% dropout, and an Adam optimizer ($1 \times 10^{-4}$ learning rate). The model outputs a binary label where '1' stands for suspicious activity, whereas '0' represents non-suspicious activities.

We obtained optimal scores on the generated summaries when utilizing the BMT+seq HOI+NMS, as observed in table~\ref{tab:ucf-bleu-table}.  Further, the NMS value used here is 0.47 (based on table~\ref{tab:nms-bleu-table}). Table~\ref{tab:mdvc-bmt-table} lists various metrics for evaluating different models. This array of metric evaluations is because not a single (standalone) metric can help with which model performs better. This behavior can be due to the objective nature of these metrics that fail to attribute to the subjective nature of the natural language generation of high-priority applications such as suspicious activity detection using summaries. Hence, the various metrics for evaluating our model against the current state-of-the-art models.

\section{Evaluation and Results}
\label{results}
This section evaluates the proposed methodology and states the results associated with this research. These results are then expanded by a comparative analysis later in Section \ref{ablative}.
\begin{table*}
\caption{A comparative analysis of BLEU and METEOR scores between generated summaries and ground truth summaries on UCF-Crime dataset for the ground truth generated.}
\label{tab:ucf-bleu-table}
\centering
\small
\begin{tabular}{c|c|c|c|c|c}
\toprule 
\textbf{Metric} & \textbf{BLEU@1} & \textbf{BLEU@2} & \textbf{BLEU@3} & \textbf{BLEU@4} & \textbf{METEOR} \\ 
\midrule
Vanilla BMT & 10.47 & 5.43 & 0.114 & 0.053 & 13.22 \\
BMT+NMS & 12.43 & 6.96 & 0.39 & 0.018 & 13.79 \\
BMT+Par HOI & 8.32 & 4.31 & 0.29 & 0.056 & 11.65 \\ 
BMT+Par HOI+NMS & 14.43 & 5.64 & 1.25 & 0.37 & 14.80 \\
\midrule
\textbf{BMT + Seq HOI + NMS} & \textbf{14.49} & \textbf{7.03} & \textbf{2.54} & \textbf{1.19} & \textbf{15.05} \\
\bottomrule
\end{tabular}
\end{table*}

\subsection{Dense Video Captioning}
\label{sec:eval_vs}
The evaluation of MDVC and BMT on ActivityNet Captions \cite{krishna2017dense} was performed by \cite{iashin2020multi} (represented in table~\ref{tab:mdvc-bmt-table}). Table~\ref{tab:ucf-bleu-table} describes these BLEU (for uni-gram (BLEU@1), bi-gram (BLEU@2), tri-gram (BLEU@3), and quad-gram (BLEU@4)) and METEOR between the output summarizations on a subset of UCF-Crime. Refer to~\ref{sec:Bleu} and ~\ref{sec:METEOR} for a detailed explanation of BLEU and METEOR scores. Table \ref{caption-table4} demonstrates a sample of summaries generated by the Vanilla BMT, BMT+Seq HOI+NMS, and BMT+Par HOI+NMS. As we discussed earlier, the DE-TR model generates a two-Stream output of Optical Flow and RGB features, as its Sampling rate is comparatively faster when compared to all other HOI models.

\subsection{Text classification}
The BERT-based cased and uncased text classifier uses Binary Classification Accuracy for evaluation. We fine-tuned the pre-trained BERT-base-uncase embeddings and model to a training accuracy of 97.29\% on the total textual corpus (divided into training and testing sets in a 7:3 ratio). The 7:3 split preserved a similar distribution for suspicious (315 for training and 135 for testing) and non-suspicious activities (330 for training and 145 for testing) as the complete dataset. This model achieved a testing accuracy of 96.84\% for this configuration of data split.

\section{Comparative Analysis}
\label{ablative}
\subsection{Non-Maximum Suppression (NMS) values} In this research, a genetic algorithm optimizes the tuning of NMS threshold values for different temporal Intersection-over-union (tIoU). The NMS values were selected for a pre-trained model and changed subsequently for each generation of parents (over five generations) based on the fitness score (here, BLEU@1). The NMS Threshold of 0.47 was achieved for Mutation parameters of [0.01, 0.03, 0.05, 0.07, 0.1] added based on a Gaussian probability distribution and randomly used for each parent over each generation with only 10\% UCF-Crime dataset on an NVIDIA A100 GPU for approximately 92 hours. Table \ref{tab:nms-bleu-table} represents the top 8 results for the exploration of NMS thresholds. Since traditional exploratory analysis can lead to a local extremum, using a Genetic Algorithm imposes a meta-heuristic on the NMS Threshold search that mitigates this problem. It is also important to note that other meta-heuristic search techniques might work even better for the same problem, but a comprehensive discussion is beyond the scope of this research.

\begin{table*}
\caption{A comparison of BLEU scores for different NMS values based on Genetic Optimization for 5 generations}
\label{tab:nms-bleu-table} 
\centering
\begin{tabular}{c c c c c c c c c}
\toprule \textbf{NMS thresholds} & 0.1 & 0.3 & 0.35 & 0.45 & \textbf{0.47 (best)} & 0.5 & 0.7 & 0.9\\ 
\midrule \textbf{BLEU@1 scores} & 12.44 & 12.57 & 13.44 & 13.12 & 13.97 & 12.41 & 12.11 & 11.17\\
\bottomrule
\end{tabular}
\end{table*}

\subsection{Evaluating on ActivityNet Captions dataset}
\label{subsec:evalActivityNet}
Table \ref{tab:mdvc-bmt-table} represents the values obtained using only a 25\% subset of the ActivityNet Captions dataset for dense captions \cite{krishna2017dense}. We observe the start and end times of any activity demarcate its corresponding caption (depicted in figure \ref{fig:workflow}) for all the models in table \ref{tab:mdvc-bmt-table}. It is important to note that the ActivityNet Captions dataset can have audio, hence the VGGish model was kept untouched during the training of the DE-TR model along with the Encoder Layer of the BMT+Seq HOI+NMS and BMT+Par HOI+NMS models.

\begin{figure*}[htp]
\begin{center}
    \includegraphics[width=0.8\textwidth]{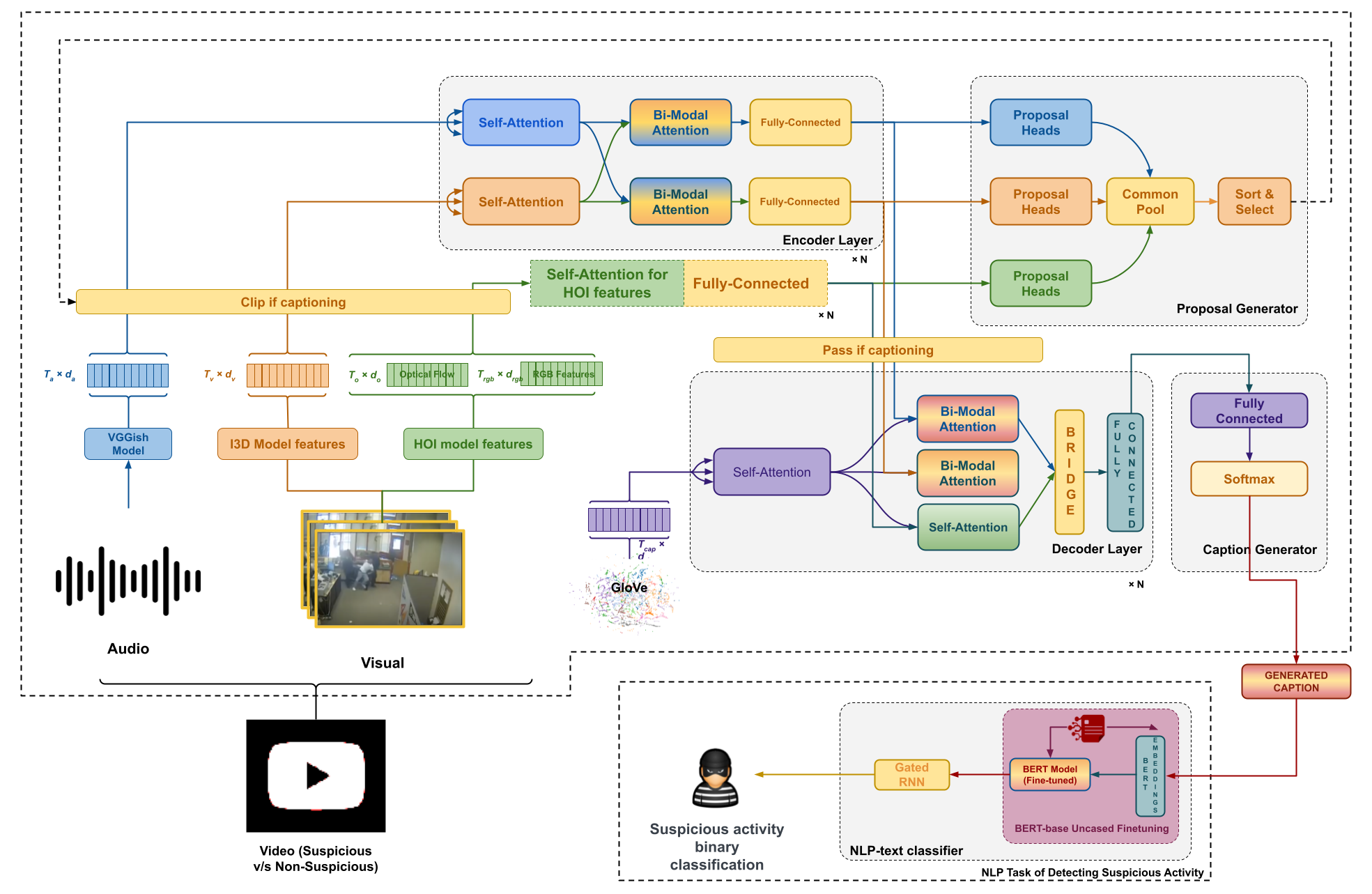}
    \caption{Representation of Model architecture of the BMT+Par HOI+NMS utilized for the Comparative Analysis \ref{ablative}.}
    \label{fig:bmtparhoi}       
\end{center}
\end{figure*}

\textbf{Comparing Sequential HOI with Parallel HOI:} We observe that the BMT + Seq HOI + NMS outperforms the BMT+Par HOI+NMS. This behavior occurs primarily because the Parallel HOI model creates an additional self-attention + proposal head pair (shown in figure \ref{fig:bmtparhoi}). This pair (self-attention + proposal head) leads to redundant feature representations that the NMS should have mitigated. Due to this, the vital feature representations vanish due to the high bias that the model learns due to redundant feature representations. But, the Sequential flow replaces the I3D model with the DE-TR (or HOI) model as the primary choice of feature extraction (discussed in the methodology). Hence, this sequential two-stream output of HOI-based Optical Flow and HOI-based RGB feature allows the model to attend to the Human-Object-Interaction feature representations more efficiently and comprehensively leverage the Proposal Generator of the BMT (by using Proposal Heads). Table \ref{tab:ucf-bleu-table} suggests that the Sequential Flow pipeline outperforms the Parallel flow pipeline in all metrics (namely, BLEU@1, BLEU@2, BLEU@3, BLEU@4, and METEOR, etc.). Table \ref{tab:mdvc-bmt-table} suggests a similar conclusion for the ActivityNet data analysis.

\begin{table*}
\caption{A comparative analysis of the models for dense captioning evaluation on ActivityNet Captions dataset. Here, only 25\% of the dataset was used for evaluation purposes.}
\label{tab:mdvc-bmt-table}
\centering
\small
\begin{tabular}{cccccc}
\toprule 
\textbf{Model} & \textbf{BLEU@1} & \textbf{BLEU@2} & \textbf{BLEU@3} & \textbf{BLEU@4} & \textbf{METEOR}\\ 
\midrule
MDVC\cite{iashin2020multi} & 12.13 & 9.34 & 4.52 & 1.98 & 11.07 \\
BMT\cite{iashin2020better} & 11.12 & 8.71 & 4.63 & 1.99 & 10.90 \\
TSP\cite{alwassel2021tsp} & 11.93 & 9.1 & 4.02 & 1.99 & 8.53 \\
PDVC (using TSP)\cite{wang2021end} & 13.84 & 9.87 & 4.47 & 2.01 & 8.94 \\
Bidirectional + Infra-Caption\cite{song2020team} & 12.71 & 10.33 & 5.11 & 2.72 & 11.28 \\
ADV-INF + Global\cite{kanani2021global} & 14.64 & 12.38 & 10.74 & \textbf{9.45} & \textbf{16.36} \\
\midrule
BMT+NMS ($\phi = 0.47$) & 13.75 & 9.71 & 4.27 & 2.01 & 12.99 \\
BMT+Par HOI+NMS ($\phi = 0.47$) & 12.43 & 8.29 & 4.26 & 1.99 & 12.47 \\
BMT+Seq HOI+NMS ($\phi = 0.47$) & \textbf{14.78} & \textbf{12.73} & \textbf{10.91} & 7.11 & 16.27 \\
\bottomrule
\end{tabular}
\end{table*}

\section{Dataset Limitations}
\label{sec:datasetlimitations}
This section contributes literature on the technical limitations of the datasets used in real-time surveillance settings (with a focus on ML and Computer Vision) based on the ActivityNet Captions and UCF-Crime datasets. We also discuss the general limitations of a suspicious activity detection model.

\begin{enumerate}[noitemsep]
\item The \textbf{size of the training and evaluation data} (in both ActivityNet and UCF-Crime) was limited to fractions of the original video dataset. The distribution of the training and evaluation set (in terms of Crime classes and length of the videos) followed the original dataset. This undersampling ensured efficient utilization of limited computational resources;
\item The \textbf{lack of an audio component} in surveillance feeds severely limits the capabilities of the dataset to work in real-time surveillance applications like crime detection;
\item The \textbf{low resolution} of real-time surveillance data can lead to overarching concerns while training Computer Vision models. For instance, in videos related to Vandalism or Arson in UCF-Crime, the activity of a person starting a fire was seldom detected by the encoder-decoder model due to how fires start under different circumstances;
\item The scarcity of multi-view stereo data sources for real-time surveillance can mitigate the problems of
\begin{itemize}
    \item Objects of Interest at a far distance. For instance, a knife in a single-view surveillance feed might not be visible if captured from farther distances;
    \item People of Interest. For example, single-view surveillance cameras may miss a high-priority crime (such as murder) due to obstructions, and hence, it might not be detected;
\end{itemize}
\end{enumerate}

\section{Ethical Considerations}
\label{sec:ethicalconsiderations}
This section discusses a few ethical considerations related to the dataset and the application discussed.
\begin{enumerate}[noitemsep]
\item Suspicious behavior is intrinsically diverse. This diversity leaves a vast scope of deliberation regarding the validation of results. Due to the unprecedented nature of the context in which the suspicious activity occurs and the possibility of a previously unknown suspicious activity, it is vital to subject the model to regular revisions and checks;
\item Both false positive and false negative results are counterproductive for real-world scenarios of detecting suspicious activities. False Positives occur when a non-suspicious activity is predicted as suspicious behavior, whereas False Negatives occur due to the prediction of suspicious activities as non-suspicious;
\item The dataset curated by the authors represents a specific sample of crimes (discussed in section \ref{data}) and is not representative of the entire population. The dataset (that the authors created) aimed at demonstrating the working of the proposed modification approach, and the results can be highly variable when deployed on a large-scale basis. 
\end{enumerate}

\section{Conclusion and Discussions}
\label{conclusion}
This paper introduced a new architecture for detecting suspicious activities in videos--by combining a Human-Object interaction model with a Dense Video Captioning model to generate Summaries. It further contributed two new datasets--ground truth captions for the UCF Crime dataset and the dataset for text classification of suspicious from non-suspicious activities.

Our work also discusses the drawbacks of using the DE-TR model in parallel for the first time and compares it with a similar Sequential contemporary for Dense Video Captioning using BMT. We achieved this by using the same model in a sequential form in a two-stream approach to replace the existing i3D features that formed the basis of the Vanilla BMT. This research also emphasized evaluating the model trained to other state-of-the-art models, and this model outperformed most of the models (represented by table~\ref{tab:mdvc-bmt-table}). We obtained these conclusive results due to many practices employed while modifying the pre-existing BMT architecture. These practices included the usage of NMS thresholding (to get a disjoint set of image sequences), setting 0-tensors (for blank audio components in videos in UCF-crime dataset while training), searching the optimal NMS threshold (by use of Genetic Algorithms), and comparing various BMT architectures.

Further, we plan to work on possible modifications to this model architecture to extract audio features that have not been utilized optimally due to the sub-optimal VGG architecture (discussed in more detail in~\cite{iashin2020better}). We further intend to compare various architectures defined in table~\ref{fig:comphoi} based on a two-stream approach with a faster sampling rate to understand the impacts on  BLEU and METEOR scores. Conclusively, we can enhance video resolution by utilizing super-resolution models to get even better summaries from the models. And we also intend to explore other hyperparameter optimization techniques for the NMS threshold and work on the Optical Flow output (from NMS+Seq HOI+BMT) by leveraging modern techniques proposed by MODETR~\cite{mohamed2021modetr}. We further intend to explore our approach for mixed-modality datasets that contain both audio and videos after leveraging domain adaptation for videos (with audio) to prevent violation of any legal terms.

\medskip


{\small
\bibliography{egbib}
\bibliographystyle{ieee_fullname}
}

\appendix

\section{Appendix}
This section describes the technical implementation of the three different models (DE-TR for HOI Detection, BMT for Video Summarization, and GRU-based BERT for Text Classification), and section~\ref{sec:supplemental} discusses the metrics leveraged for this study. describes the metrics leveraged for this study. All the model implementations were in PyTorch and heavily utilized code from GitHub repositories of the origin work.

\subsection{DE-TR for HOI Detection}
\label{sec:DETRforHOI}
As mentioned earlier, we trained the pre-trained DE-TR (or Detection Transformer model) on a subset of HICO-DET (\~19\% of the dataset) for approximately 52 hours on 2 A100 GPUs. It had a Sampling Rate of 35 fps for inference when on NVIDIA GTX 1060. The two-stream output (Optical Flow and RGB) that we tuned it for is the primary cause for this extreme compute consumption. We set a learning rate of $1e-4$, batch size of 2, weight decay of $1e-4$, and backbone learning rate of $1e-4$ for 100 epochs with early stopping of 10 for this implementation. Further, we used the default backbone of ResNet-50, and two workers were used (as the number of GPUs was 2).

\subsection{BMT for Video-Summarization}
\label{sec:BMTforVS}
We trained the BMT in a disjoint way. First, we kept the VGGish model weights the same (used in the official GitHub implementation\footnote{https://github.com/v-iashin/BMT (accessed: May 8, 2022)}). We first fine-tuned the whole Caption Generator (Encoder-Decoder) with DE-TR (discussed earlier in~\ref{sec:DETRforHOI}), as the authors suggested in the GitHub implementation (for i3D), and then fine-tuned the Proposal Generator. For training the Caption Generator, we used a subset of the UCF-Crime dataset (as discussed earlier) for 62 hours with a learning rate of $1e-4$ and a batch size of 16. We used start-times and end-times as labels for the proposals that the authors generated (using a combination of MDVC and BMT~\cite{iashin2020multi, iashin2020better}). These proposal times were matched with each sentence of our annotated data points and then used to train the Proposal Generator. We used the same learning rate and a batch size of 8 for fine-tuning the proposal generator. Here, we added a condition to change the noisy or random audio signals to 0-tensors. Note that these steps stayed the same for BMT+Seq HOI+NMS (figure~\ref{fig:architecture}) and BMT+Par HOI+NMS (figure~\ref{fig:bmtparhoi}), apart from the modifications in the model architecture. We did not use the 0-audio-tensors while evaluating the models on ActivityNet Captions dataset. Hence, the results are comparable to the Global ADV-INF model in table~\ref{tab:mdvc-bmt-table}.

\subsection{GRU-based BERT for Text Classification}
\label{sec:GRUforTC}
We trained GRU-based BERT based on the dataset discussed in section~\ref{data}. Here, we used a combination of 408 summaries modified from CHARADES (according to earlier details) and 212 from USC DPS reports. This dataset was then augmented with 300 text summaries from the UCF-Crime dataset (250 Suspicious and 50 Non-Suspicious), leading to a total of 920 data points (445 Suspicious and 475 Non-suspicious). This combined dataset was split in a 7:3 split for training and inferencing for our model. To not learn any biases from the UCF-Crime Dataset, we kept most of it in the testing set (276 summaries: 230 Suspicious and 46 Non-suspicious), and the rest contributed to the training set (24 summaries: 20 Suspicious and 4 Non-suspicious). We used Adam with a learning rate of $1e-4$ and a dropout of 0.15 with 256 hidden layer units in GRUs. The training accuracy obtained was 97.29\% and 96.84\%.

\begin{figure*}[tp]
\begin{center}
    \includegraphics[width=0.8\textwidth]{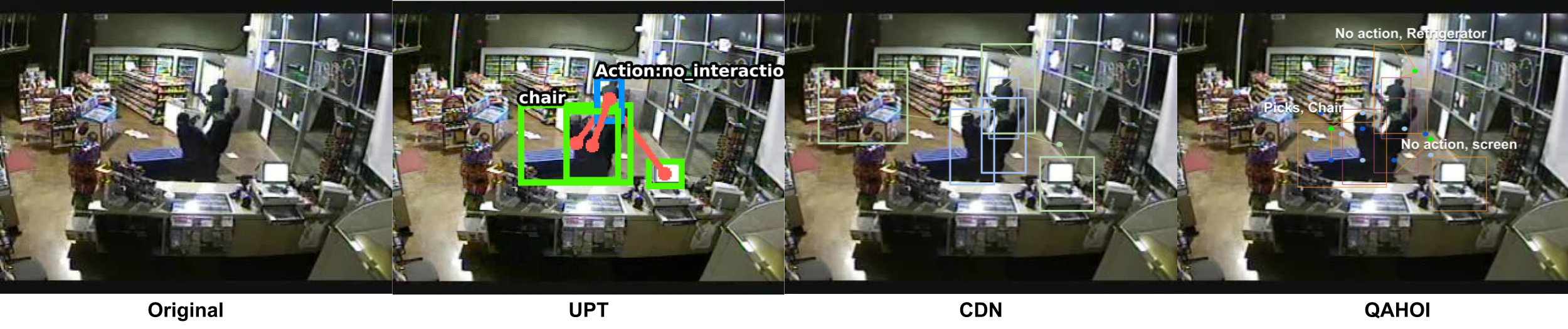}
    \caption{A comparison drawn from Unary Pairwise Transformer (ResNet 50), CDN-small, and QAHOI-Tiny models. All these models were trained on HICO-DET and have been evaluated in table~\ref{tab:comparetable}.}
    \label{fig:comphoi}       
\end{center}
\end{figure*}

You can find a comparison of the model outputs in figure~\ref{fig:comphoi}. Table~\ref{caption-table4} depicts a subsample of the model summaries generated by the Vanilla BMT model, BMT+Seq HOI+NMS, BMT+Par HOI+NMS, and the ground truth. These results suggest that the captions generated using BMT+Seq HOI+NMS follow the metric scores obtained. Further, this approach also showcases the redundancy that the Parallel HOI model introduces in the generated summaries (refer to table~\ref{caption-table4}).

\begin{table*}
\centering
\scriptsize
\begin{tabular}{l|p{0.2\linewidth}|p{0.2\linewidth}|p{0.2\linewidth}|p{0.2\linewidth}}
\toprule
\toprule
\textbf{Video class} & \textbf{Vanilla BMT} & \textbf{BMT + Par HOI} & \textbf{BMT + Seq HOI}  & \textbf{Ground Truth} \\ 
\midrule
Normal & \scriptsize{A man is working on a car . He continues to talk about the car and the man continues to talk . A person is seen walking around a car and speaking to the camera . A man is seen walking down a street and driving down a street . He then takes a tire off the car and continues to talk . We see a title screen . He then takes the car off the car and continues to wash the car . We see the ending title screen . We see a black opening screen.} & \scriptsize{We see the ending title screen. The man then walks away and walks away. He gets up and gets up and gets up and gets up and gets up. He gets up and walks away.} & \scriptsize{A group of people are standing on a table. A boy is seen sitting on a table with a toy boy standing on his hands and leads into a boy standing on a boy holding a toy. A group of people are standing on a table. The video ends with a man walking away from the camera. A group of people are standing on a table.} & \scriptsize{A man comes into the frame and then leaves the video frame. Other customers also leave. The counter is packes with man and woman. The grey shirt woman leaves the store. The man leaves the counter.} \\
Robbery & \scriptsize{A man is working in a room . A man is working on a house . A man is seen standing in a room holding a box and speaking to the camera .} & \scriptsize{A man is seen speaking to the camera while holding a camera and leads into him holding a bottle of beer. A man is seen sitting on a table with a knife and holding a knife. He then takes a drink and puts it down on the table. A man is seen sitting on a table with a dart board and looking at the camera. A person is seen sitting on a table with a white cloth.} & \scriptsize{A \textbf{man holding gun} is seen talking and leads into him walking around him. A \scriptsize{man holding gun} is seen standing in a room and leads into him walking up a door. A man holding a bag is seen standing in a room with a dog in a house.} & \scriptsize{A group of men enters a jewellery shop. One of the men breaks all the glass showcases containing the jewelleries. Then, the men start taking all the jewelleries from the showcases and put those in their bags.Finally, they leave the shop with their bags.}\\
Assault & \scriptsize{A man is walking in a room . A man is working on a bike . The man is in the room , he is wearing a black shirt and he is wearing a black shirt and he is wearing a black shirt and he is . A man is sitting on a chair . A man is seen walking up to the camera while another man walks away . A man is sitting on a stairs . A man is sitting on a couch} & \scriptsize{A man is standing in a room talking to the camera . A man is seen walking around a room while another man is walking around the room . A man is working on a wall . A man is seen walking around a room while another man is walking around and walking around . A man is working on a roof} & \scriptsize{A man in a black shirt is standing in front of a woman in front of a woman. A man in a black shirt is hit in front of a woman and falls down. A man in a black shirt is standing in front of a woman in front of a woman. A man in a black shirt is standing in front of a woman in front of a woman.} & \scriptsize{A lot of people are standing on the pavement in front of a building. Suddenly, some people start hitting a man who finally faints and falls on the ground.}\\
\bottomrule
\bottomrule
\vspace{0.02cm}
\end{tabular}
\caption{Some samples of summaries generated by the models and ground truths (generated by the authors).}
\label{caption-table4}
\end{table*}

\section{Supplemental Material}
\label{sec:supplemental}

\subsection{Non-Maximum Suppression (NMS) and time-Intersection-over-Union (tIOU)} 
\label{sec:NMS}
The objects in the image can be of different sizes and shapes, and to capture each of these perfectly, the object detection algorithms create multiple bounding boxes. NMS is a common technique in Object Detection (used in this case) where it selects the few best bounding boxes based on a threshold. In other words, non-max suppression (or non-maximum suppression or NMS) helps to find the best bounding box for an object and reject or `suppress' all other bounding boxes.

In other words, NMS is a technique to select one entity out of many overlapping entities. As we know, various frames in a video can have a high degree of overlap, so NMS is employed to select frames with a lower degree (only here) of overlap. Hence, NMS helps segment frames unrelated to each other and leads to better results when decoding feature representations for generating captions (due to better clipping obtained as in figure~\ref{fig:architecture}). This NMS threshold takes two things into account.

\begin{enumerate}
\item Objectiveness Score (given by the model)
\item Overlap or IOU of the bounding boxes (for Human and Object Detection) on a frame-by-frame basis
\end{enumerate}

\begin{figure*}[tp]
\begin{center}
    \includegraphics[width=\textwidth]{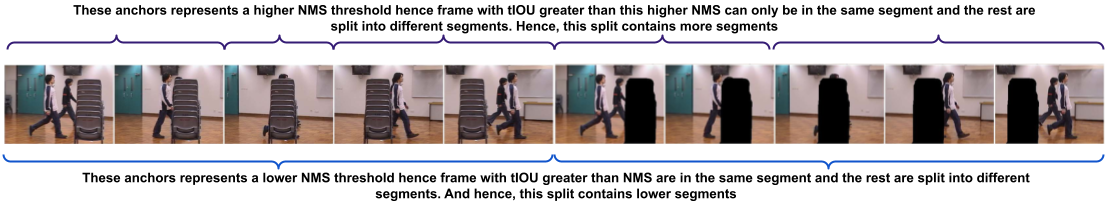}
    \caption{Representing tIoU (time-Intersection-over-Union) and NMS (Non-Maximum Suppression) frame-by-frame from a video from CHAIRS~\cite{jia2004video}. Here, the Blue anchors define a segmentation of frames in the video where corresponding frame-pairs have higher NMS threshold. With lower NMS threshold, less number of segments contain several frames while with a higher NMS threshold, more number of segments contain less number of frames}
    \label{fig:NMSfortime}       
\end{center}
\end{figure*}

A clear definition of NMS is represented in figure~\ref{fig:NMSfortime}, along with the anchors to understand the concept. The aim of the Genetic Algorithm Optimization strategy was to achieve a balance between the number of frames (in segments) and segments (N; for which the number of repeated blocks of Encoder-Decoder in architecture). Since most videos are encoded, the BMT model aims to obtain RGB and Optical Flow features. An NMS threshold addresses the RGB and Optical Flow features obtained to get better segments of frames for better captions.

Hence, we define a term–tIOU (time-Intersection-over-Union), which measures the overlap between the frames and tries to keep unrelated Frames in different segments and similar ones in one Segment. When defining an NMS threshold, we are querying all the Segments with an NMS threshold higher than a given value. On the contrary, Frame pairs with lower tIoU get divided into different Segments. This segregation results in different Segments in text summaries.

\subsection{Average and Mean Average Precision}
\label{sec:mAP}
The Average Precision (or AP) is the most popular evaluation metric for object detection and human-object interaction models. HOI tasks deal with a range of Intersection-over-Union (IoU) threshold values (different from tIOU discussed in~\ref{sec:NMS}). AP is the weighted mean of precision values at these IoU thresholds. On the other hand, Mean Average Precision (mAP) is the average of the Average precision values for each output class and is the area under the Precision-Recall curve. This metric is closely related to the concepts of Confusion Matrix, Intersection-over-Union, Precision, and Recall. To calculate the Average Precision, the following are the steps.\\

\begin{enumerate}
    \item Generate the prediction scores using the model
    \item Convert the prediction scores to class labels
    \item Calculate the confusion matrix—TP, FP, TN, FN
    \item Calculate the precision and recall metrics
    \item Calculate the area under the precision-recall curve
    \item Measure the average precision
\end{enumerate}

And after taking the mean of the calculated AP, we obtain the mAP (Mean Average Precision) value discussed earlier.

\subsection{BLEU score} 
\label{sec:Bleu}

BLEU (Bilingual Evaluation Understudy) is a metric for evaluating machine-translated pieces of text. It has values between 0 and 1 that score the similarity of a machine-translated text to some high-quality reference translated texts. A higher BLEU score implies better translations or summarizations. Here, the BLEU score is a percentage of values from 100 instead of a value from 1. Further, the values (for instance, BLEU@n refers to the BLEU score for individual BLEU values for n-grams) in the official definition of BLEU are not a single value of BLEU but a whole family of them parametrized by the weighting vector. In other words, it is the weighted Geometric Mean of the modified n-gram precision. The BLEU score (eq.~\ref{eq:1}) is the product of the weighted GM and the brevity penalty (penalizing short text generation to contain words in the original/ground-truth corpus).

\begin{equation}
\label{eq:1}
BP = e^{-(r/c-1)^{+}}
\end{equation}

Here, $(r/c-1)^{+}$ refers to the positive part of this term. $r$ is the effective length of the ground-truth corpus. Hence, when $r<c$, no penalization occurs, and BP is 1.\\

\subsection{METEOR Score} 
\label{sec:METEOR}
METEOR is another metric for evaluating machine-translated texts based on the harmonic mean of the Precision and Recall of Unigrams, where Precision values are weighted more than Recall values. Similar to BLEU, higher METEOR scores imply better machine translations. This metric can also account for the problems in the BLEU metric. For instance, Word stems and synonyms are not handled well in BLEU (discussed above). This metric is of significant importance as it has a higher correlation with human-readable generated texts. Eq.~\ref{eq:2} is used to calculate this metric.

\begin{equation}
\label{eq:2}
M = F_{hmwp} \cdot (1 - p) 
\end{equation}

Here, M is the METEOR score, $F_{hmwp}$ is the harmonic mean in which Precision ($P$) is weighted ten times more than the Recall ($R$), and has been depicted by eq.~\ref{eq:3}. $p$ designates the penalty incurred when considering longer text segments. This penalty is higher if more mappings are incongruent with Segments of texts in the original (/expected) text summaries. This penalty is given by eq.~\ref{eq:4}.

\begin{equation}
\label{eq:3}
    F_{hmwp} = \frac{10\cdot P\cdot R}{R + 9\cdot P}
\end{equation}

Here, $P$ is Precision, and $R$ is the Recall of the model used.

\begin{equation}
\label{eq:4}
    F_{hmwp} = 0.5 \cdot (\frac{c}{ u_{m}})^{3}
\end{equation}

Here, $u_{m}$ is the number of mapped unigrams, and $c$ is the number of segments mapped for these.


\end{document}